# REALIZATION OF ONTOLOGY WEB SEARCH ENGINE

Olegs Verhodubs

oleg.verhodub@inbox.lv

**Abstract.** This paper describes the realization of the Ontology Web Search Engine. The Ontology Web Search Engine is realizable as independent project and as a part of other projects. The main purpose of this paper is to present the Ontology Web Search Engine realization details as the part of the Semantic Web Expert System and to present the results of the Ontology Web Search Engine functioning. It is expected that the Semantic Web Expert System will be able to process ontologies from the Web, generate rules from these ontologies and develop its knowledge base.

I. INTRODUCTION

The rapid growth of the Web shows the law of the passage of quantitative changes into qualitative changes in action. The development of the Web has provided us with more information than we can comprehend or manage effectively [1]. Keyword-based search engines such as YAHOO, YANDEX, GOOGLE and others marked the first qualitative leap of the Web. They provied users with links to relevant pages in the Web. At first, there was an adequate response to the ever-increasing Web. Over time, the use of keyword-based search engines is becoming less effective owing to some difficulties. Firstly, relevant pages, retrieved by search engines, are useless, if they are distributed among a large number of mildly relevant or irrelevant pages. Secondly, relevant pages may not be retrieved by search engines at all. In truth, it is a rare phenomenon that happens with modern search engines. Thirdly, the pages, retrieved by search engines, are sensitive to vocabulary. Initial keywords do not provide with the results we want, because relevant pages use different terminology from the original query [2]. Finally, retrieved results are single web pages and hence it is necessary manually extract the partial information and put it together [3]. Additional work is needed to process retrieved pages for extracting necessary information. Considering ever-increasing scope of the Web, the difficulties will exacerbated in the future. Apparently, it is time to the the second qualitative leap of the Web.

Actually, the potential for qualitative changes had been accumulated, and the changes were occuring at the moment. More precisely, the direction of the Web future development has been crystallized, but detailed road map towards the desired result was not worked out to the end. There is a consensus that the Web should become more intelligent, however the foundation of the intelligence continues to be developed. It is fraught with the possibility of variations. Hope to use an existing representation in the Web for more advanced processing of the text was not justified. There was nothing better to reconsider the nature of information representation in the Web. The Semantic Web technologies were developed.

Ontologies are one of the most important parts of the Semantic Web technologies. They give an opportunity to represent information in the Web in more machine processable way. OWL (Web Ontology Language) is a Semantic Web language to represent ontologies in the Web [4]. Having been developed, it was expected that the Semantic Web Expert System would generate rules from OWL ontologies, supplement its knowledge base with generated rules and to reason, based on these rules and user interaction [5]. The task of rule generation from OWL ontologies had already been solved [6]. In turn, it is necessary for the Semantic Web Expert System to have an ability to search OWL ontologies in the Web according to the user request. Ontology Web Search Engine was a project that incarnated the ability of ontology search in the Web and that has already been described in theory [2]. The main purpose of this paper is to describe the realization of the Ontology Web Search Engine and also to demonstrate its functioning. The Ontology Web Search Engine functioning has been evaluated by means of testing, and it is reflected in the paper, too.

This paper is divided into several sections. The next section presents the structure of the Ontology Web Search Engine. The third section sheds light on the Ontology Web Search Engine realization details and testing. The last section is dedicated to the conclusions of the work and to the plans for future research.

II. ONTOLOGY WEB SEARCH ENGINE STRUCTURE

The task of the ontology search had required its decision a long time ago. As was known, necessity gave birth to proposals, and the solution of the ontology search was found soon. One of such proposals identified the steps of the ontology search and also defined the input as a set of keywords and defined the output as a list of ontologies [7]. This was the most used scheme of the semantic search that was realized many times in the form of the Semantic Web search engines. SWOOGLE was one of the Semantic Web search engines for indexing and retrieving the Semantic Web documents including RDF (Resource Description Framework) and OWL [8]. SWOOGLE is realized as a web page, and it is available at http://swoogle.umbc.edu. WATSON was one more search engine for the Semantic Web [9]. This search engine is available at http://watson.kmi.open.ac.uk/WatsonWUI/. There are a lot of other Semantic Web search engines as Falcons, Sindice, Semantic Web Search, SWSE (Semantic Web Search Engine), but it makes no sense to overviw all of them, because they are very similar.

There are two alternatives for realizing the Ontology Web Search Engine [2]. The first alternative inolves the creation of a complete web search engine like SWOOGLE or WATSON. The second alternative involves the creation of a web search engine as a part of another system. We are interested in the second alternative, and another system in our case is a Semantic Web Expert System.

The typical structure of a web search engine comprises four essential modules [10]:

- crawler,
- an indexer,
- a query engine,
- a page repository.

Crawlers are computer programs that browse the Web [10]. Crawlers use a starting set of URLs (Uniform Resource Locator) and retrieve the content on the Web pages specified by the URLs. The crawlers extract URLs appearing in the retrieved web pages and visit some or all of these URLs, thus repeating the retrieval process.

The indexer takes all the words from each document in the page repository and records the URL, where each word occurred [10]. The result is a very large database, which provides the URLs that point to pages, where a given word occurs. The database may also contain other structural information such as links between documents, incoming URLs to these documents, formatting aspects of the documents, and location of terms with respect to other terms .

The web query engine receives the search requests from users [10]. It takes the query submitted by the user, splits the query into terms, and also searches these terms in the database, which is built by the indexer. The web query engine then retrieves the documents that match the terms within the query and returns these documents to the user .

The page repository stores the Web content that was retrieved during the crawling process [10].

Semantic Web search engines are very similar to traditional web search engines, and the main difference between them is that the Semantic Web search engines work not only with documents in HTML, but also with documents in RDF and OWL. Considering also that there is no need to create a complete and self-sufficient system for searching ontologies, it is necessary to implement three separate computer programs. One of these programs is a crawler for searching OWL ontologies in the Web. The crawler has to browse the Web, extract URLs appearing in the retrieved web pages, visit these URLs if they are URLs to HTML web pages and store these URLs if they are URLs to RDF or OWL documents. One more computer program is an indexer for indexing or storing data about each OWL ontology. The data about indexed ontologies is stored in a page repository. The indexer stores all the information as ontology URL and constructs about each ontology in the page repository. The result is a very large database, which provides all the information about ontologies. The last computer program is a query engine. This program is necessary to search ontology in the page repository according to the user's request, and it is embeded in the Semantic Web Expert System.

These three programs are separate programs, and consequently they run separately, too. The order of starting of these programs is very important [2]. The crawler should be started first. The indexer should be started after the crawler finishes its work. The query engine can be started after the crawler and the query engine finishes their work. In the next section the realization of the Ontology Web Search Engine and its testing are described.

III. ONTOLOGY WEB SEARCH ENGINE REALIZATION AND TESTING

A: Realization

The implementation of the Ontology Web Search Engine in the form of three separate programs was done using a computer with Intel® Celeron® Processor B800 1.5 GHz and 2 GB RAM. Jbuilder 2008 was chosen as an Integrated Development Environment. Java programming language was selected due to the use of the Apache Jena in the Semantic Web Expert System. Jbuilder was selected due to personal preferences.

Implementing of the crawler, it was decided to use one of the existing crawlers. There are a lot of existing crawlers. They are Heritrix, Jspider, Crawler4j, Nutch and many other crawlers. Certainly, each of these crawlers had its own advantages and disadvantages, but Crawler4j was chosen for use [11]. The main requirements for the crawler are that it would be realized in Java, would be open-source and would be well-documented. Crawler4j has several parameters. One of them is the first URL, from which the crawler starts to browse the Web [11]. One more is the maximum depth of crawling. The next one is the maximum number of pages to fetch. It is also possible to set the number of crawlers that is the number of concurrent threads for crawling and to set the politeness that is the period of time between requests. It necessary to mention that this crawler was transformed in order to store URLs of ontologies. Ontologies are the files with the .rdf and .owl extentions, but URLs of ontologies are stored in the usual text file. This file is used by the indexer to access the ontologies for indexing.

Implementing of the ontology indexer, it was decided to use one of the existing indexers. There are such indexers as Lucene, Nutch, Solr, Sphinx and others. The main requirements for implementing of the indexer are the same as the requirements for the crawler that is it would be realized in Java, would be open-source and would be well-documented. Apache Lucene was chosen [12]. It was also necessary to use some tool for processing ontologies to realize the indexer. This is necessary to extract such ontology components as classes, properties and relations. For this purpose Apache Jena was used [13]. Apache Jena has its own disadvantages as an inability to work with all formats of OWL ontology, however the choice of Apache Jena is due to the fact that the Semantic Web Expert System has already exploited

Apache Jena in its other subsystems. Apache Lucene creates an index on the disk, which can be used by the query engine. The index is a folder with several files. This folder has to be copied to the place, where it will be used. This means that it is necessary to copy the folder to the place, where the query engine is placed.

Implementing of the query engine, Apache Lucene was used, too. Apache Lucene was used not only for indexing of data, but also for retrieveing of the indexed data [12]. For this purpose the query engine used the index, created in the process of indexing.

B: Testing

The crawler was implemented first. The crawler was tested varying its parameters, while at the same time some parameters remained constant. There are constant parameters such as the first URL, the maximum depth of crawling and the politeness. The first URL is set to http://www.ontologyportal.org. The maximum depth of crawling is set to -1 that means unlimited depth of crawling. The parameter of the politeness is set to 300 that means each request after 300 milliseconds. Thus, the number inner of crawlers and the number of web pages to fetch are varied in the process of testing. We are interested in the number of found ontologies and in the time spent on searching ontologies. Table I shows the results of the crawler testing.

TABLE I. Characteristics of the crawler functioning.

| Number of inner crawlers | Number of web page to fetch | Number of found ontologies | Time of searching, ms | First URL |
|---|---|---|---|---|
| 1 | 500 | 191 | 1532227 | http://www.ontologyportal.org |
| 1 | 1000 | 195 | 2276391 | http://www.ontologyportal.org |
| 1 | 3000 | 207 | 7756870 | http://www.ontologyportal.org |
| 1 | 5000 | 246 | 10827814 | http://www.ontologyportal.org |
| 1 | 7000 | 327 | 16605796 | http://www.ontologyportal.org |
| 2 | 500 | 191 | 981577 | http://www.ontologyportal.org |
| 2 | 1000 | 210 | 1270897 | http://www.ontologyportal.org |
| 2 | 3000 | 239 | 3623858 | http://www.ontologyportal.org |
| 2 | 5000 | 269 | 6690904 | http://www.ontologyportal.org |
| 2 | 7000 | 340 | 7917062 | http://www.ontologyportal.org |
| 3 | 500 | 191 | 962010 | http://www.ontologyportal.org |
| 3 | 1000 | 191 | 863058 | http://www.ontologyportal.org |
| 3 | 3000 | 234 | 2732466 | http://www.ontologyportal.org |
| 3 | 5000 | 266 | 4782998 | http://www.ontologyportal.org |
| 3 | 7000 | 273 | 5892416 | http://www.ontologyportal.org |
| 4 | 500 | 190 | 261810 | http://www.ontologyportal.org |

| 4 | 1000 | 190 | 441932 | http://www.ontologyportal.org |
| 4 | 3000 | 196 | 1402221 | http://www.ontologyportal.org |
| 4 | 5000 | 239 | 3279246 | http://www.ontologyportal.org |
| 4 | 7000 | 241 | 3925284 | http://www.ontologyportal.org |

Analyzing the table II, it is possible to come to conclusion that the most effective functionig mode of the crawler is provided with the parameters of two inner crawlers and the 7000 number of web pages to be fetched, because in this case the greatest number of ontologies are found. It is possible to state that if the number of ontologies to be fetched would be greater than 7000, the number of found ontologies woud be greater, too. The main parameter here is the number of used inner crawlers, which equals to two. It is difficult to say why the use of two inner crawlers is more effective.

The indexer was the next implemented program. It was tested by means of indexing of 1035 found ontologies (more precisely ontology URLs). The URL's of these ontologies were accumulated using the crawler several times and starting it from different beginning URLs. The process of ontology indexing was going on during 68 minutes. The index, created by the indexer, held 2,15 Mb space on disk, however only 262 ontologies were indexed. This is due to a lot of null URLs, plenty of empty ontologies that is ontologies without classes, properties, relations, due to ontologies in the format, which is not supported by Apache Jena and also due to huge size of some ontologies (more than 3 Mb). The indexer was programmed do not index ontologies, which held more than 3 Mb space on disk, because of weak computer use.

The query engine was implemented last. Testing of the query engine showed that the index was created qualitatively. The query engine displayed the ontology URLs according to the typed keywords. If typed keyword equals to the name of class, property or relation, an appropriate ontology URL is displayed.

IV. CONCLUSION

This paper demonstrates the realization and testing results of the Ontology Web Search Engine as the part of the Semantic Web Expert System. It is obvious that the realization of the Ontology Web Search Engine as a separate and professional system requires the improvement of all Ontology Web Search Engine parts. For example, the crawler and the indexer have to work in parallel if we speak about professional Ontology Web Search Engine. Found ontology URLs have to be stored in the database, but not in the text file. It is possible that the crawler has to be implemented in programming language C++ in order to work faster. In turn, there is a need for indexing of all ontologies, regardless of the format of ontologies. In this connection, it is necessary to utilize one more framework for ontology processing in addition to Apache Jena.

As for the general conclusions, ontologies cannot displace regular web pages completely at this moment. It was convinced once more in the process of testing the crawler, since this conclusion was presented in the previous paper [2]. Indeed, 1035 ontologies of different quality were found, only. Of course, it is possible to state that there are more than 1035 ontologies in the Web, but this quantity is much less than quantity of regular Web pages. For example, this may be changed like this. An effective algorithm of ontology generation from the web page has to be placed to each web server so that each web server distributive has to have the possibility of ontology generation from the web page. Thus, the web page developer realizes a web page as before, but a web server will generate the ontology, based on the web page automatically if the corresponding option is enabled. So, the algorithm of ontology generation from the web page is really needed. The task of ontology generation from the web page and also some other possibilities of knowledge extraction from the plain text are the next fields of research that will be worked out within the Semantic Web Expert System.